\documentstyle[colap]{article}

\title{\vspace{-0.5in} Focus and Higher--Order Unification} 
\author{Claire Gardent \\
Computational Linguistics \\
Universit\"at des Saarlandes, \\
D--66041 Saarbr\"{u}cken \\
{\tt claire@coli.uni-sb.de} 
\And
Michael Kohlhase \\
Computer Science \\ 
Universit\"at des Saarlandes, \\
D--66041 Saarbr\"{u}cken \\
{\tt kohlhase@cs.uni-sb.de}}

\newcommand{\bc}{\begin{center}}
\newcommand {\ec}{ \end{center}}
\newcommand {\be} {\begin{enumerate}}
\newcommand {\ee} {\end{enumerate}}
\newcommand {\bi} {\begin{itemize}}
\newcommand {\ei} {\end{itemize}}
\newcommand {\ba} {\vspace*{2pt}\noindent$\begin{array}}
\newcommand {\et} {\end{tabular}\vspace*{2pt}}
\newcommand {\bt} {\vspace*{2pt}\noindent\begin{tabular}}
\newcommand {\ea} {\end{array}$\vspace*{2pt}}
\newcommand {\txt}[1]{\mbox{{ \it{#1}}}}
\def\wff{\txt{wff}}
\newcommand {\ra}{\rightarrow}

\def\Covars{{\cal C\kern-.2ex V}}

\def\fsv#1{\overline{#1}}
\newtheorem{defn}{Definition}[section]

\begin{document}

\maketitle
\vspace{-0.5in}
\begin{abstract}
Pulman has shown that Higher--Order Unification (HOU) can
be used to model the interpretation of focus. In this
paper, we extend the unification--based approach to cases which are
often seen as a test--bed for focus theory: utterances with multiple
focus operators and second occurrence expressions. We then show that the
resulting analysis favourably compares with two prominent theories of
focus (namely, Rooth's Alternative Semantics and Krifka's Structured
Meanings theory) in that it correctly generates interpretations which
these alternative theories cannot yield.  Finally, we discuss the
formal properties of the approach and argue that even though HOU need
not terminate, for the class of unification--problems dealt with in
this paper, HOU avoids this shortcoming and is in fact computationally
tractable.
\end{abstract}

\section{Introduction}

In this paper, we argue that Higher--Order Unification (HOU) provides
a linguistically adequate tool for modeling the semantics of focus.
Building up on~\cite{Pulman:houatiof95}, we develop a
unification--based analysis of focus which we show favourably compares
with two prominent theories of focus, Rooth's Alternative Semantics
and Krifka's Structured Meanings theory.  For data which is generally
viewed as a test--bed for focus theory (utterances with multiple focus
operators and second occurrence expressions), we show that contrary to
Rooth's and Krifka's theories, the HOU treatment yields a transparent
analysis while avoiding under-- and over--generation.

\section{Focus theory}
\label{s2}

Focus is a much debated notion. In this paper, we assume a simplified
version of Jackendoff's definition: a {\bf focus} is the semantic
value of a prosodically prominent element. We take the
identification of prosodically prominent elements as given.

To set the stage for this paper, we will briefly review the folklore,
i.e. the main issues of focus theory. It is commonly agreed that focus
triggers the formation of an additional semantic value which we will
call the {\bf Focus Semantic Value} (FSV).  The name and definition of
the FSV varies from author to author:
Jackendoff~\cite{Jackendoff:siigg72} calls it the {\it
  presuppositional set}, Rooth~\cite{Rooth:atofi92} the {\it
  Alternative Set} and Krifka~\cite{Krifka:acsfmfc92} the {\it
  Ground}. In this paper, we assume a definition of the FSV which is
in essence Rooth's Alternative set, that is, the set of semantic
objects obtained by making an appropriate substitution in the focus
position. For instance, the FSV of (1a) is defined as (1b), the
set of properties of the form {\it like--ing y} where $y$ is an
individual (in what follows, focus is indicated using upper--case; we
also follow Montague's convention that for any type $\tau$, $D_{\tau}$
is the set of objects of type $\tau$ and \wff$_{\tau}$ is the set of
wffs of type $\tau$).

\bt{lll}
(1) & a. & {\it Jon only likes MARY} \\
& b. & $\{ \lambda x. l(x,y) \mid y \in D_e \} $
\et

It is also usually agreed that certain linguistic elements {\bf
  associate with focus} in that the meaning of the utterance
containing these elements varies depending on the choice of focus. For
instance in (2a--b), the focus operator {\it only} associates with
focus so that the difference in focus between (2a) and (2b) induces a
difference in meaning between the two utterances: in a world where Jon
introduced Paul to Mary and Sarah, and no other introduction takes
place, (2a) is necessarily false whilst (2b) is true.

\bt{lll}
(2) & a. & {\it Jon only introduced Paul to  MARY} \\
& b. & {\it Jon only introduced PAUL to  Mary}
\et

To model this ``association--with--focus" phenomenon, the semantics of
associating--elements (e.g. focus operators, quantificational adverbs)
is made contingent on the FSV which itself, varies with the choice of
focus. The following example illustrates this. Suppose that the
meaning of {\it only} is determined by the following rule: 

\bt{l}
[NP only VP] \\
$\hookrightarrow \forall P [P \in FSV \wedge P(NP') \ra P = VP']$ 
\et

where $NP', VP'$ represent the meaning of NP and VP respectively, and
$FSV$ stands for the focus semantic value of the VP. As we have seen
above, the FSV of (1a) is (1b), hence by the above semantic for {\it
  only}, the semantics of (1a) is:

\bt{l}
 $\forall P [P \in \{ \lambda x. l(x,y) \mid y \in D_e \}
\wedge P(j)$ \\
 $\phantom{\forall P [} \ra P = \lambda x. l(x,m)]$
\et

Intuitively, the only property
of the form {\it like--ing y} that holds of {\it Jon} is the property
of {\it like--ing Mary}. 

\section{The basic analysis}
\label{s3}

For computing the Focus Semantic Value, we propose to use
Higher--Order Unification. More specifically, given (part of) an
utterance $U$ with semantic representation $Sem$ and foci $F^1 \dots
F^n$, we require that the following equation, the {\bf ground
  equation}, be solved:

\ba{l}
 Sem = Gd(F^1) \dots (F^n)
\ea

Assuming the typed $\lambda$--calculus as our semantic representation
language, this equation can be solved by Huet's algorithm (cf.
\cite{Huet:auaftlc75}), thus assigning a value to $Gd$. On the basis of this
value, we can then define the FSV, written $\fsv{Gd}$, as follows:

\begin{defn}(Focus Semantic Value) \\
  Let Gd be of type $\alpha =\vec{\beta_{k}}\ra t$ and $n$ be the
  number of foci ($n\leq k$), then the Focus Semantic Value derivable
  from Gd, written $\fsv{Gd}$, is $\{Gd(t^1\dots t^n )\mid
  t^i\in\wff_{\beta_i }\}.$
\end{defn}

As mentioned before, this yields a focus semantic value which is in
essence Rooth's Alternative Set\footnote{Though in fact, our
  definition is more syntactic than Rooth. In Rooth's approach, the
  FSV definition is purely semantic whereas in our approach the FSV is
  indirectly defined by solving equations and the value thus obtained
  (i.e. the value of $Gd$) is a {\it term}, that is, a syntactic
  object. Hence, our FSV can be more accurately compared to Kratzer's
  {\it presupposition skeleton},.  This means that our approach
  inherits the advantages of Kratzer's approach (cf.
  \cite{Kratzer:trof91}). In particular, it adequately captures the
  interaction of focus with VP ellipsis as illustrated by Kratzer's
  notorious example: {\it I only went to TANGLEWOOD because you
    did}.}.

Finally, we assume as in \cite{Pulman:houatiof95}, that foci are stored and discharged non--deterministically as the need arises, thus contributing to the definition of the ground equation. Furthermore, equations are set up at the level at which there are needed e.g. at the VP level in the case of a pre--verbal focus operator. 

To illustrate the workings of our approach, we now run through a
simple example. Consider (1a). To determine the meaning of {\it only
  likes MARY}, the FSV of the VP must be known. Hence the
following equation must be solved:

\ba{l}
\lambda x. l(x,m) = Gd(m)
\ea

By HOU, the value of $Gd$ is then\footnote{\label{fn3} Unification
  yields another possible value of $Gd$, namely $ \lambda y \lambda x.
  l(x,m)$. In what follows, we assume a restriction similar to the
  DSP's {\bf Primary Occurrence Restriction}~\cite{DaShPe:eahou91}'s:
  the occurrence directly associated with the focus is a primary
  occurrence and any solution containing a primary occurrence is
  discarded as linguistically invalid. For instance, $m$ is a primary
  occurrence in the equation $\lambda x. l(x,m) = Gd(m)$ so that the
  solution $ Gd = \lambda y \lambda x. l(x,m)$ is invalid. For a
  formal treatment of DSP's Primary Occurrence Restriction and a
  discussion of how it can be extended to focus, see
  \cite{GaKo:hocuanls96}.}:

\ba{l}
Gd = \lambda y \lambda x. l(x,y) 
\ea

And by definition (3.1), the FSV is:

\ba{l}
\fsv{Gd} =  \{ \lambda x. l(x,y) \mid y \in \wff_e \}
\ea

Assuming the semantic of {\it only} given above, the semantic
representation of (1a) is then:

\ba{l}
\forall P [P \in \{ \lambda x. l(x,y) \mid y \in \wff_e \}
\wedge P(j)\\
\phantom{\forall P [} \ra P = \lambda x. l(x,m)]
\ea

In short, we obtain a reading similar to that of Rooth, the difference
being in the way the FSV is determined: by HOU in our approach, by
means of a semantic definition in Rooth's.

\section{Linguistic applications}
\label{s4}

In this section, we show that the HOU approach favourably compares
with Rooth's and Krifka's analysis in that it correctly generates
interpretations which these two theories fail to yield. As we
shall see, the main reason for this is that the HOU approach makes
minimal assumptions about the role syntax plays in determining the
FSV. In particular, it relies neither on the use of Quantifier
Raising, nor on the assumption of a rule--to--rule definition of
the FSV. In this way, it avoids some of the pitfalls these
theories encounter.

We begin by a brief summary of Rooth's and Krifka's theories and
stress the properties relevant for the present discussion.
We then confront the three theories with the data.

\subsection{Two alternative theories of focus}
\label{s41}

\subsubsection*{Rooth's Alternative Semantics}

In Rooth's approach, the FSV is defined by recursion on the
truth--conditional structure which is itself derived from LF
(i.e. Logical Form, the Government and Binding level of semantic
representation). Focus is then seen as introducing a free variable
whose value is determined by the current context and is furthermore
constrained to be an element or a subset of the FSV. For our purpose,
the following characteristics are particularly important:

\bi
\item Given Rooth's definition of the Alternative Set, a focus
  operator associates with any focus occurring in its
  scope.
\item Any NP may be subject to Quantifier Raising. Importantly, this includes focused NPs.
\item Quantifier Raising may not apply to quantifiers occurring in a scope--island.
\ei

Note that Rooth's approach critically relies on {\it quantifier raising} as
a means of moving a focused NP out of the scope of a focus operator.
However this only applies if the focus NP is not embedded in a {\it scope
island}.

\subsubsection*{Krifka's Structured Meanings}

Krifka's approach defines a rule--to--rule semantics which assigns to
any syntactic constituent, a meaning which can be either a
$\lambda$--term or a structured meaning, i.e. a tuple of the
form $\langle Gd, F \rangle$ where $Gd$ is Krifka's
Focus Semantic Value and $F$ is a (possibly complex) focus. 

For our purpose, an important characteristic of Krifka's approach is
the tight syntax/semantic interaction it presupposes. In particular,
the theory requires that a focus operator combines with a syntactic
constituent $C$ whose structured semantics $C' = \langle Gd, F
\rangle$ provides the focus ($F$) this operator associates with. In
other words, the right--adjacent sibling of a focus operator must
contain all and only the foci this operator associates with.  As we
shall later see, some of the data does not seem to square with this
assumption.

\subsection{Multiple Focus Operators}
\label{s42}

Utterances with multiple focus operators\footnote{The subscripts
  indicates which operators associate with which focus. There are
  there for clarity only, and have no theoretical import.} are known
pathological cases of focus theory:

\bt{lll}
(3) & a. & {\it (Jon only$_1$ read the letters} \\
&&{\it that Sarah sent to
  PAUL$_1$)} \\ 
 & b. & {\it Jon also$_2$ only$_1$ read the letters} \\
&&{\it that SUE$_2$ sent to PAUL$_1$.}
\et

In the given context, the preferred reading of (3b) can be glossed as
follows: {\it it is also the case for SUE$_2$, that Jon only$_1$ read the
  letters she sent to PAUL$_1$ -- i.e. Jon didn't read the letters she$_2$
  sent to e.g.  Peter}. In other words, the preferred reading is that
{\it also$_2$} associates with {\it SUE$_2$} and {\it only$_1$} with {\it
  PAUL$_1$}.

\subsubsection*{The HOU analysis}

Under the HOU approach, (3b) is analysed as follows. First, the
meaning of {\it only$_1$ read the letters that SUE$_2$ sent to
  PAUL$_1$} is derived. To determine the FSV of the VP, the ground
equation (4b) must be solved for which (4c) is a solution. Applying
the semantics of {\it only} given in section \ref{s2}, the semantics
of (4a) is then as given in (4d)\footnote{For clarity, we have
  simplified the semantic representation of (3b); nothing hinges on
  this.}.

\ba{lll}
(4) & a. &  \txt{only$_1$ read the letters that SUE$_2$} \\
&& \txt{sent to PAUL$_1$} \\
& b. & G^{1}(p) = \lambda x . read(x,l(s,p)) \\
& c. & 
G^{1} = \lambda y. \lambda x . read(x,l(s,y)) \\
& d. & 
\lambda z. \forall P [P \in  \fsv{\lambda y \lambda x
  . read(x,l(s,y))} \wedge P(z)  
\\
&&\phantom{\lambda z. \forall P [}\ra P = \lambda x. read(x,l(s,p)) ] 
\ea

Analysis then proceeds further and the ground equation 
 \[ G^{2}(s) = \lambda z. \forall P [
\begin{array}[t]{l}
   P \in  \fsv{\lambda y \lambda x . read(x,l(s,y))}\\
    \wedge P(z)\ra P = \lambda x. read(x,l(s,p)) ]
   \end{array}\]
must be solved to determine the meaning of {\it also$_2$ only$_1$ read the letters that
  SUE$_2$ sent to PAUL$_1$}. A possible solution for $G^{2}$ is 
\[ \lambda u. \lambda x . 
   \lambda z. \forall P[
   \begin{array}[t]{l}
    P \in  \fsv{\lambda y \lambda x . read(x,l(u,y))}\\
     \wedge P(z)\ra P = \lambda x. read(x,l(u,p)) ]
   \end{array}\]
   Assuming the  following semantics for $\txt{NP also VP}$
\[\exists P[P \in \txt{FSV} \wedge P(NP') \wedge P\neq VP']\]
we obtain the desired reading
\[\exists P[%
\begin{array}[t]{l}
P \in  \fsv{\lambda u \lambda x .
   \begin{array}[t]{l}
          \txt{only$_1$ read the letters that}\\
          \fsv{\txt{u sent to Paul$_1$}}
   \end{array}}\\
  \wedge P(j)\wedge P \neq\lambda z. z
    \begin{array}[t]{l}
       \txt{only$_1$ read the letters}\\
       \txt{that Sue$_2$ sent to Paul$_1$}]
    \end{array}
\end{array}\]

\subsubsection*{Comparison with Rooth and Krifka}

As mentioned in section \ref{s41}, under the Alternative Semantics
approach, a focus operator necessarily associates with any focus
occurring in its scope. Furthermore in (3b), the scope of {\it
  only$_1$} is the whole VP {\it read the letters that SUE$_2$ sent to
  PAUL$_1$}.  Hence, if no quantifier raising occurs, {\it only$_1$}
associates with both {\it SUE$_2$} and {\it PAUL$_1$}. Thus in order
to generate the desired reading, {\it SUE$_2$} must be moved out of
the scope of {\it only$_1$}. However, since the NP {\it the letters
  that SUE$_2$ sent to PAUL$_1$} is a scope island, quantifier raising
is impossible. Hence, the desired reading cannot be
generated\footnote{This point is independently noted in
  \cite{Rooth:atofi92}.}.

Recall that in the Structured Meanings approach, the right--sibling of
a focus operator must contain all and only the focus this operator
associates with (cf. section \ref{s41}). Hence, to generate the
desired reading in (3b), there must exist a syntactic constituent
which is right--adjacent to {\it only$_1$} and which contains {\it PAUL$_1$}
but not {\it SUE$_2$}\footnote{This is a simplification: the constituent
  may in fact contain {\it SUE$_2$} but this focused NP should already
  have been bound by some focus operator so that the focus of the
  whole constituent only includes {\it PAUL$_1$}. Since no focus operators
  occur in this constituent, it follows that such constituent does not
  exist.}; similarly, there must exist a syntactic constituent which
is right--adjacent to {\it also} and which contains {\it SUE$_2$} but not
{\it PAUL$_1$}.  Given standard assumptions about syntax, such
constituents do not exist so that the desired interpretation cannot be
generated.

\subsection{Second Occurrence Expressions}
\label{s43}

We call second occurrence expressions (SOE) utterances which partially
or completely repeat a previous utterance. Typical cases of SOEs are:
corrections (5a), echo--sentences (5b) and variants (5c).

\bt{llll}
(5) & a. & A: &  {\it Jon only likes MARY.} \\
& & B: &   {\it No, PETER only likes Mary.} \\
& b. & A: &   {\it Jon only likes MARY.} \\
& & B: &   {\it Huhu, Peter only likes Mary.} \\
& c. & A: &   {\it Jon only likes MARY.} \\
& & B: &   {\it So what? Even PETER only}\\
& &    &  {\it likes Mary.} 
\et

An important property of SOEs is that the repeated material is {\bf
  deaccented}, that is, it is characterised by an important reduction
in pitch, amplitude and duration (cf. \cite{Bartels:sot95}).  On the other
hand, all three theories of focus considered here are based on the
assumption that focus is prosodically marked and thus, identifiable.
Hence, the question arises of whether these theories can account for
SOEs.

\subsubsection*{The HOU analysis}

Our proposal is to analyse SOEs as involving a deaccented anaphor
which consists of the repeated material, and is subject to the
condition that its semantic representation must unify with the
semantic representation of its antecedent.

This is modeled as follows. Let $SSem$ and $TSem$ be the semantic
representation of the source (i.e. antecedent) and target (i.e.
anaphoric) clause respectively, and $TP^1 \dots TP^n$, $SP^1 \dots
SP^n$ be the target and source parallel elements\footnote{As in
  \cite{DaShPe:eahou91}, we take the identification of parallel
  elements as given -- for the moment.}, then the interpretation of an
SOE must respect the following equations:

\ba{l}
An(SP^1, \ldots ,SP^n) = SSem \\
An(TP^1, \ldots ,TP^n) = TSem
\ea

Intuitively, these two equations require that target and source clause
share a common semantics $An$, the semantics of the deaccented
anaphor.

Given this proposal, the analysis of (5a) involves three equations:

\ba{l}
An(j) = \forall P [P \in  \fsv{\lambda y \lambda x. l(x,y)}\\
\phantom{An(j) =\forall P [} \wedge P(j)\ra P = \lambda x. l(x,m) ]  \\
An(p) =\forall P [P \in \fsv{Gd} \wedge P(p)\ra P = \lambda x. l(x,m) ] \\ 
Gd(F) = \lambda x. l(x,m) 
\ea

Since neither $Gd$ nor $Focus$ are initially given, the third equation
above is untyped and cannot be solved by Huet's algorithm\footnote{Even
  though this is not explicitly stated, Pulman's analysis~\cite[page
  6]{Pulman:houatiof95} faces a similar problem.}. In that situation,
we can either assume some delaying mechanism or some extension of
Huet's algorithm that can cope with type variables (cf.
\cite{Dougherty:houuc93,Hustadt:actsfphou91}).  Resolution of the
first equation yields the following solution:

\ba{l}
An = \lambda y \forall P [P \in \{ \lambda x. l(x,y) \mid y \in
\wff_e \} \\
\phantom{An = \lambda y \forall P [} \wedge P(z) \ra P = \lambda x. l(x,m) ] 
\ea

By applying $An$ to $p$, the left--hand side of the second equation is
then determined so that the second equation becomes  

\ba{l}
\phantom{=}   \forall P [P \in  \fsv{\lambda y \lambda x. l(x,y)}\wedge P(p) \ra P = \lambda x. l(x,m) ]\\
=  \forall P [P \in \fsv{Gd} \wedge P(p)\ra P = \lambda x. like(x,m) ]
\ea

 and the value of $Gd$ is identified as being
\[Gd = \lambda y \lambda x. l(x,y)\]
(Note further, that the third equation can now be solved thus yielding
the value $m$ for the focus $F$.)  That is, the HOU approach to
SOEs allows us to correctly capture that fact that an SOE can inherit
its FSV from its source clause (by unification). In
\cite{GaKoLe:cahou96}, we show in more detail how the analysis accounts
for the interaction of focus with anaphora and definiteness in the
case of a particular instantiation of SOEs, namely corrections.

\subsubsection*{Comparison with Rooth and Krifka}

Under the Alternative Semantics approach, SOEs are captured as
follows. It is assumed that the quantification domain of focus
operators is a variable whose value is contextually determined. In the
standard case (i.e. the case where the focus is prosodically marked),
this quantification domain of focus operators is usually identified
with the FSV of the VP.  However, in the SOE cases, the assumption is
that the quantification domain of focus operators is identified with
the FSV of the source clause. Thus in (5a), the quantification of
{\it only} in the second clause is identified with the FSV of the
preceding utterance i.e. the set of properties of the form {\it
  like--ing somebody}.

But now, consider the following example:

\bt{lll}
(6) & a. & {\it Jon only likes MARY.} \\
& b. & {\it * No, PETER only likes Sarah.} 
\et

Clearly, this dialog is ill--formed in that (6b) is no appropriate
correction for (6a). However, under the Alternative Semantics
approach, it will not be ruled out since the FSV of (6a) provides an
appropriate quantification domain for the focus operator in (6b): as
required by the semantic of pre--verbal {\it only}, it is a set of
properties whose elements can be identified with the VP semantic value
$\lambda x. l(x,m)$. Hence although Rooth's approach captures some
cases of SOEs, it does not seem to provide an adequate
characterisation of the phenomena at hand.

The Structured Meanings proposal dis\-tin\-gui\-shes be\-tween {\it proper--}
and {\it quasi--}SOEs. Proper--SOEs involve an exact repetition of
some previous linguistic material, and are analysed as involving an
anaphor which is constrained by the restriction that it be a segmental
copy of its antecedent.  For instance, the semantics of {\it only
  likes Mary} in (5b) is not determined by the semantics of its parts
but is instead identified with the semantic value of its antecedent
{\it only likes MARY} in (5a). In contrast, quasi--SOEs only involve
{\it semantic} equivalence between repeating and repeated material
(for instance, in a quasi--SOE a repeated element may be
pronominalised). Krifka claims that quasi--SOEs have prosodically
marked foci and thus do not raise any specific difficulty.

However this theory faces a number of methodological and empirical
difficulties. First, it is non--compositional because the meaning of
the deaccented material in proper-SOEs is solely defined by the
meaning of its antecedent (rather than the meaning of its parts).
Second, the prosodic data is rather unclear: the assumption that
quasi--SOE contains a prosodically marked focus is a moot point (cf.
\cite{Bartels:sot95}) and if it proves to be false, the analysis fails
to account for quasi--SOEs. Third, it is counterintuitive in that it
handles separately two classes of data (i.e.  quasi-- and
proper--SOEs) which naturally belong together. Indeed, the HOU
approach can be shown to provide a uniform treatment of quasi-- and
proper--SOEs (cf.~\cite{GaKoLe:cahou96}).

\section{Formal properties of the HOU approach}
\label{s5}
The unification problem can be stated as follows: Given two terms of a
logic {\bf M} and {\bf N}, is there a substitution, $\sigma$, of terms
for variables that will make the two terms identical (i.e.
$\sigma$({\bf M}) = $\sigma$({\bf N}))?

It is well-known that for Higher--Order Logic (e.g. the typed
$\lambda$--calculus) the space of solutions can be infinite and
furthermore, the HOU problem is only semi--decidable so that the
unification algorithm need not terminate for unsolvable problems.

Fortunately, in our case we are not interested in general unification,
but we can use the fact that our formulae belong to very restricted
syntactic subclasses, for which much better results are known. In
particular, the fact that free variables only occur on the left hand
side of our equations reduces the problem of finding solutions to
higher-order matching, of which decidability has been proven for the
subclass of third-order formulae~\cite{Dowek:tomid92} and is
conjectured for the general case. This class, (intuitively allowing
only nesting functions as arguments up to depth two) covers all of our
examples in this paper.  For a discussion of other subclasses of
formulae, where higher-order unification is computationally feasible
see~\cite{Prehofer:dhoup94}.

\section{Conclusion}
\label{s6}

In this paper, we have argued that Higher--Order Unification provides
an adequate tool for computing Focus Semantic Values. To this end, we
have considered data which is viewed as a test--bed for focus theory
and shown that, whilst existing theories either under--generate,
over--generate or are methodologically unsatisfactory, the HOU
approach yields a simple and transparent analysis. There appear to be
two main reasons for this.

First, the HOU analysis makes minimal assumptions about the role
syntax is called to play in determining the FSV. It is defined on a
purely semantic level in the sense that unification operates on
semantic representations, and relies neither on quantifier raising,
nor on a rule-to-rule definition of the FSV. As we have seen, this
type of approach is a plausible way to avoid under--generation.

Second, the HOU approach permits an equational analysis which can
naturally be further constrained by additional equations. The interest
of such an approach was illustrated in our treatment of SOEs which we
characterise as involving two phenomena: the computation of an FSV,
and the resolution of a deaccented anaphor. Not only did we show that
this analysis is methodologically and empirically sound, we also
showed that it finds a natural realisation in the equational framework
of HOU: each linguistic phenomena is characterised by some equation(s)
and the equations may mutually constrain each other. For instance, in
the case of SOEs, we saw that the equations characterising the
deaccented anaphor help determine the unidentified FSV of the
utterance containing the unmarked focus.

Clearly, our approach extends to cases of adverbial quantification.
For lack of space we could not develop the theory here; let us just
point out that von Fintel's criticism~\cite{Fintel:amtoaq95} of
semantic approaches to focus, also applies to Krifka's Structured
Meanings analysis, but not to the HOU approach presented here. Von
Fintel points out that in certain cases of adverbial quantification, a
focus operator associates with an {\it unmarked} focus and does {\it
  not} associate with a marked focus occurring in its scope -- as
should be clear from this article, this is unproblematic for our
analysis.

Of course, there are still many open issues. First, how does the
proposed analysis interact with quantification? Second, how does it
extend to a dynamic semantics (e.g. Discourse Representation Theory)?

\section{Acknowledgments}

The work reported in this paper was funded by the Deutsche
Forschungsgemeinschaft (DFG) in Sonderforschungsbereich SFB--378,
Project C2 (LISA).

%\bibliographystyle{acl}
%\bibliography{coling96}

\end{document}